\documentclass[conference]{IEEEtran}
\IEEEoverridecommandlockouts
\usepackage{cite}
\usepackage{amsmath,amssymb,amsfonts}
\usepackage{algorithmic}
\usepackage{graphicx}
\usepackage{textcomp}
\usepackage{xcolor}
\usepackage{algorithm2e}%
\usepackage{booktabs}
\usepackage{acronym}
\newacro{OCC}[OCC]{One-Class Classification}

\def\BibTeX{{\rm B\kern-.05em{\sc i\kern-.025em b}\kern-.08em
    T\kern-.1667em\lower.7ex\hbox{E}\kern-.125emX}}
\begin{document}

\title{Newton Method-based Subspace Support Vector Data Description}

\author{

\IEEEauthorblockN{Fahad Sohrab}
\IEEEauthorblockA{\textit{Faculty of Information Technology}\\
\textit{and Communication Sciences} \\
\textit{ Tampere University}\\
Tampere, Finland \\
 fahad.sohrab@tuni.fi}
\and 
\IEEEauthorblockN{Firas Laakom}
\IEEEauthorblockA{\textit{Faculty of Information Technology}\\ \textit{and Communication Sciences} \\
\textit{ Tampere University}\\
Tampere, Finland \\
 firas.laakom@tuni.fi}

\and
\IEEEauthorblockN{Moncef Gabbouj}
\IEEEauthorblockA{\textit{Faculty of Information Technology}\\ \textit{and Communication Sciences} \\
\textit{ Tampere University}\\
Tampere, Finland \\
moncef.gabbouj@tuni.fi}
}

\maketitle

\begin{abstract}
In this paper, we present an adaptation of Newton's method for the optimization of Subspace Support Vector Data Description (S-SVDD). The objective of S-SVDD is to map the original data to a subspace optimized for one-class classification, and the iterative optimization process of data mapping and description in S-SVDD relies on gradient descent. However, gradient descent only utilizes first-order information, which may lead to suboptimal results. To address this limitation, we leverage Newton's method to enhance data mapping and data description for an improved optimization of subspace learning-based one-class classification. By incorporating this auxiliary information, Newton's method offers a more efficient strategy for subspace learning in one-class classification as compared to gradient-based optimization. The paper discusses the limitations of gradient descent and the advantages of using Newton's method in subspace learning for one-class classification tasks. We provide both linear and nonlinear formulations of Newton's method-based optimization for S-SVDD. In our experiments, we explored both the minimization and maximization strategies of the objective. The results demonstrate that the proposed optimization strategy outperforms the gradient-based S-SVDD in most cases. 
\end{abstract}


\section{INTRODUCTION}
Traditional machine learning binary classification tasks focus on developing models that accurately classify samples into two categories. However, when faced with imbalanced datasets, where one category has a significantly larger number of samples than the other, machine learning models tend to exhibit bias towards the majority category. This bias can lead to suboptimal performance. To address this issue, one-class classification techniques can be used. These techniques involve training a model using only data from one category rather than both categories. This can help to mitigate bias and improve the model's ability to classify samples from the minority category.

\ac{OCC} approaches are a set of unsupervised learning techniques, typically carried out by using instances in the target class. The trained model is used to distinguish the target class from the rest of all classes, referred to as outliers. Considerable research has been carried out over the last few decades on learning algorithms for \ac{OCC}, and these techniques have been used for various applications. In \cite{pantazi2019automated}, \ac{OCC} is used for automated leaf disease detection in different crop species. In \cite{guo2011tumor}, \ac{OCC} technique is used to detect tumors in brain CT images. In \cite{malik2022support}, a fault detection technique for grid-connected photovoltaic Inverters based on \ac{OCC} is proposed. In \cite{sohrab2020boosting}, the capability of \ac{OCC} to complement deep CNN-based taxa identification by indicating samples potentially belonging to the rare classes of interest for human inspection is analyzed. In \cite{degerli2022early}, uni-modal and multi-modal one-class classification techniques are examined for Early Myocardial Infarction Detection. More recently, in \cite{kilickaya2023hyperspectral}, \ac{OCC} is used for hyperspectral image analysis.

\ac{OCC} methods can be broadly divided into support vector-based and non-support vector-based approaches \cite{sohrab2020ellipsoidal}. The support vector-based methods identify the so-called support vectors in the training set and are used to infer the decision boundary. The non-support vector-based approaches, such as density estimation and reconstruction-based methods, are used when the data's distribution or data generation process is accurately known. While most of the traditional algorithms for \ac{OCC} such as One-class Support Vector Machine (OC-SVM) \cite{scholkopfu1999sv} and Support Vector Data Description (SVDD) \cite{tax2004support} operate for data points in the original feature space, there has been a rising trend of algorithms where the method maps the original data to a lower dimensional feature space more suitable for one-class classification. 

Subspace Support vector data description (S-SVDD) \cite{sohrab2018subspace} is one such example of \ac{OCC} methods where the model maps the data to a subspace optimized for one-class classification. The iterative optimization process of data mapping and the data description relies on gradient descent (GD). However, GD only relies on first-order information to find its solution, which can be sub-optimal \cite{nocedal1999numerical,guler2010foundations}. In this work, we propose a novel variant of S-SVDD, where the data mapping and parameters are iteratively optimized using Newton's method rather than gradient descent. In addition to first-order information, Newton's method uses second-order information, i.e., Hessian matrix, to optimize the objective. 

 Recently, there has been an interest in using second-order optimization methods in general and Newton's method in particular in the machine learning context \cite{xu2020second,sra2012optimization,jain2017non,castera2021inertial}, as these methods have several desired theoretical properties \cite{guler2010foundations} and have shown superior empirical performance in several tasks \cite{zhong2007regularized,rafati2020quasi,lin2007trust,xu2020second,berahas2022quasi}. One major limitation of GD is sensitivity to scaling and linear transformations, which can lead to being trapped in trivial solutions \cite{guler2010foundations}. On the other hand, Newton's method does not have this limitation and is affine invariant. Furthermore, Newton's method has, in general, faster convergence rates compared to GD and yields superior performance in practice \cite{xu2020second,zhong2007regularized,zhang2019gradient,rafati2020quasi,castera2021inertial}. 

 Moreover, as the S-SVDD optimization problem involves primarily quadratic terms, computing such quantities can be done with a minor additional computational cost, which makes Newton's method a more suitable approach for this task. In this paper, we provide a more efficient strategy to solve S-SVDD based on Newton's method. In the proposed method, we compute the Hessian matrix and use it in the optimization process, which leads to accommodating the intrinsic curvature of the function in the iterative process.

\section{Newton's method-based subspace support vector data description}\label{svddexplain}

\begin{figure*}[ht]
	\centering
	\includegraphics[scale=0.65]{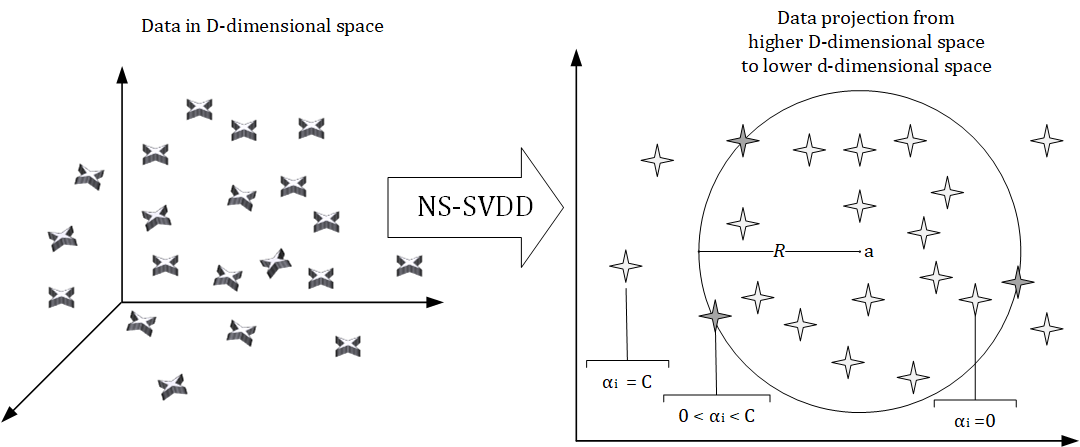}
	\caption{An illustration of data transformation from a high-dimensional space to a lower-dimensional space, optimized for one-class classification.}
	\label{approach}
\end{figure*}

Let's consider a set of data points denoted by as a matrix $\mathbf{X}=[\mathbf{x}_{1},\mathbf{x}_{2},\dots \mathbf{x}_{N}],\mathbf{x}_{i} \in \mathbb{R}^{D}$, where $N$ denoted the number of instances, and each instance $\mathbf{x}_i$ is represented in a feature space of dimensionality $D$. The goal is to find an optimized projection matrix $\mathbf{Q} \in \mathbb{R}^{d \times D}$ for the projection of the data from $D$-dimensional space to a lower $d$-dimensional space such that the data description for one-class classification in the lower $d$-dimensional space can be obtained better. The data in the lower $d$-dimensional space is represented as
\begin{equation}\label{eq:Y_i}
\mathbf{y}_i = \mathbf{Q} \mathbf{x}_i, \:\:, i=1,\dots,N,
\end{equation}

To encapsulate the given training data inside a closed boundary, a hypersphere is constructed around the data. The hypersphere is characterized by center $\mathbf{a}$ and radius $R$. The goal is to minimize the volume of the hypersphere under the constraints that most of the training data should lie inside the hypersphere, i.e.:

\begin{align}\label{erfunc}
\min \quad F(R,\mathbf{a}) = R^2 \nonumber\\
\textrm{s.t.} \quad  \| {\mathbf{Q}\mathbf{x}_i} - \mathbf{a} \|^2_{2} \le R^2, \:\: i=1,\dots,N,
\end{align} 
The slack variables $\xi_i,\:i=1,\dots, N$ are introduced to \eqref{erfunc} for accommodating outliers in the given training set. The optimization problem now becomes
\begin{align}\label{ssvdd}
\min \quad F(R,\mathbf{a}) = R^2 + C\sum_{i=1}^{N} \xi_i \nonumber\\
\textrm{s.t.} \quad  \|\mathbf{Qx}_i - \mathbf{a}\|_2^2 \le R^2 + \xi _i,\nonumber\\
\xi_i \ge 0, \:\: \forall i\in\{1,\dots,N\},
\end{align}  
Where $C>0$ is the hyperparameter used for controlling the trade between the number of instances outside the hypersphere and the volume of the hypersphere. The constraints are incorporated into the objective function by using Lagrange multipliers.

\begin{eqnarray}\label{lang}
L= {R^2} + C\sum_{i=1}^{N} \xi _i - \sum_{i=1}^{N} \gamma_i \xi_i 
-\sum_{i=1}^{N} \alpha_i \Big( R^2 + \xi _i - \nonumber\\ \mathbf{x}_i^\intercal \mathbf{Q}^\intercal \mathbf{Q} \mathbf{x}_i + 2\mathbf{a}^\intercal \mathbf{Q} \mathbf{x}_i- \mathbf{a}^\intercal\mathbf{a} \Big),  
\end{eqnarray}
with Lagrange multipliers $\alpha_i\ge0$ and $\gamma_i\ge0$. By setting the partial derivative to zero, we get:
\begin{eqnarray}
\frac{\partial L}{\partial R}=0 &\Rightarrow& \sum_{i=1}^{N} \alpha_i = 1 \label{der1} \\
\frac{\partial L}{\partial \mathbf{a}}=0 &\Rightarrow& \mathbf{a} = \sum_{i=1}^{N} \alpha_i \mathbf{Q}\mathbf{x}_i \label{der2} \\
\frac{\partial L}{\partial \xi _i}=0 &\Rightarrow& C- \alpha _i - \gamma _i  = 0 \label{der3}
\end{eqnarray} 
 Substituting \eqref{der1}-\eqref{der3} into \eqref{lang}, and denoting the data in lower dimensional space according to \eqref{eq:Y_i} we get
\begin{equation}\label{Lang2}
L = \sum_{i=1}^{N} \alpha _i \mathbf{y}_i^\intercal  \mathbf{y}_i - \sum_{i=1}^{N}\sum_{j=1}^{N} \alpha _i \mathbf{y}_i^\intercal \mathbf{y}_j \alpha _j.
\end{equation}  
Solving \eqref{Lang2} is equivalent to solving SVDD \cite{tax2004support} in the lower $d$-dimensional space. Maximizing \eqref{Lang2} will give us $\alpha$ values for all training instances. 
\subsection{Regularization}
Analogous to S-SVDD, after determining the optimal set of $\alpha_i, \:i=1,\dots, N$, we optimize an augmented version of the Lagrangian function. We add a regularization term $\Psi$, expressing the class variance in the lower d-dimensional space to \eqref{Lang2}. Hence, \eqref{Lang2} now becomes as follows
\begin{equation}\label{lang3}
L= \sum_{i=1}^{N} \alpha _i  \mathbf{x}_i^\intercal \mathbf{Q}^\intercal \mathbf{Q} \mathbf{x}_i - \sum_{i=1}^{N}\sum_{j=1}^{N} \alpha_i \mathbf{x}_i^\intercal \mathbf{Q}^\intercal \mathbf{Q} \mathbf{x}_j \alpha_j + \beta\Psi,
\end{equation}
where, the hyperparameter $\beta$ is used to control the importance of $\Psi$. The regularization term is defined as follows
\begin{equation}
\label{generalconstraint}
\Psi = Tr(\mathbf{Q}\mathbf{X} \lambda \lambda^\intercal \mathbf{X}^\intercal\mathbf{Q}^\intercal).
\end{equation} 

Where $Tr(\cdot)$ is the trace operator, the vector $\lambda \in \mathbb{R}^{N}$ is utilized to control the impact of individual samples in the regularization term. The regularization term can take three forms based on the $\lambda$ vector. In the first form, denoted as $\psi1$, all training data points contribute equally to the regularization. This is achieved by replacing all elements of the $\lambda$ vector with 1. In $\psi2$, the lambda vector is replaced by the corresponding $\alpha$ values of the instances. In $\psi3$, only the alpha values of the samples corresponding to the class boundary are replaced by the corresponding $\alpha$ and zero for other instances. When no regularization term is used in the optimization, we refer to that case as $\psi0$.
\subsection{Newton's Method-based update}
 We update the elements of the projection matrix $\mathbf{Q}$ iteratively using Newton's method as follows. First the Projection matrix is vectorized as $\mathcal{Q}=\text{Vec}(\mathbf{Q})$, which has the following form:
\begin{equation}\label{vecQ}
\mathcal{Q}^\intercal=[{Q}_{11}, {Q}_{12}, ..., {Q}_{1D},{Q}_{21}...,{Q}_{dD}],
\end{equation} 
The vectorized form of the projection matrix is updated as follows:
\begin{equation}\label{newtonupdate}
 \mathcal{Q} \leftarrow \mathcal{Q} - \eta [(H_L)^{-1}\Delta \mathcal{L}],
\end{equation} 
where $\eta$ is the learning rate used in the optimization process and $\Delta\mathcal{L}=\text{Vec}(\Delta L)$ has the following form
\begin{equation}\label{vectorizeDeltaL}
\Delta \mathcal{L}^\intercal=[\Delta L_{11}, \Delta L_{12}, ..., \Delta L_{1D},\Delta L_{21}...,\Delta L_{dD}].
\end{equation}
The elements of \eqref{vectorizeDeltaL} are computed from the gradient matrix of \eqref{lang3}. The gradient $\Delta L$ is determined by taking the derivative of \eqref{lang3} with respect to the projection matrix $\mathbf{Q}$ as follows:
\begin{eqnarray}\label{gradL}
\Delta L =2\mathbf{Q}\mathbf{X}\mathbf{A}\mathbf{X}^\intercal-   2\mathbf{Q}\mathbf{X}\alpha\alpha^\intercal \mathbf{X}^\intercal
+ \beta\Delta \Psi,
\end{eqnarray}  
where $\Delta \Psi$ in \eqref{gradL} denotes the gradient of regularization term which is computed as
\begin{equation}\label{deltapsi1}
\Delta \Psi = 2\mathbf{Q} \mathbf{X} \lambda \lambda^\intercal \mathbf{X}^\intercal.
\end{equation}  

$H_L$ in \eqref{newtonupdate} is the hessian matrix,
  
  $$
  H_L=
\begin{vmatrix}

\frac{d^2L}{d\mathcal{Q}_1\mathcal{Q}_1}& \cdots &\frac{d^2L}{\mathcal{Q}_1\mathcal{Q}_n}& \cdots &\frac{d^2L}{\mathcal{Q}_1\mathcal{Q}_{dD}}\\
\vspace{-5mm} . &&.&&.\\
\vspace{-5mm}  . &&.&&.\\
 . &&.&&.\\
\frac{d^2L}{d\mathcal{Q}_m\mathcal{Q}_1}& \cdots &\frac{d^2L}{\mathcal{Q}_m\mathcal{Q}_n}&\cdots &\frac{d^2L}{\mathcal{Q}_m\mathcal{Q}_{dD}}\\
\vspace{-5mm} . &&.&&.\\
\vspace{-5mm}  . &&.&&.\\
 . &&.&&.\\
\frac{d^2L}{d\mathcal{Q}_{dD}\mathcal{Q}_1}& \cdots &\frac{d^2L}{\mathcal{Q}_N\mathcal{Q}_n}& \cdots &\frac{d^2L}{\mathcal{Q}_{dD}\mathcal{Q}_{dD}}\\
\end{vmatrix}.
$$
In order to compute the elements of the $H_L$, we need to compute the second-order partial derivatives accordingly. We start with the first-order derivative using identities 133 and 134 in \cite{matrixcookbook}:
\begin{equation} \label{genrule}
\frac{\partial L}{\partial Q_{ij}}=tr\Big[ \big[\frac{\partial L}{\partial \mathbf{Q}} \big]^\intercal  \mathbf{S}^{ij}\Big],
\end{equation}
where $\mathbf{S}^{ij}$ is referred to as the structure matrix with the single-entry, 1 at $(i, j)$ and zero elsewhere. Putting \eqref{gradL} in \eqref{genrule}, we get

\begin{eqnarray} \label{der1H}
\frac{\partial L}{\partial Q_{ij}}=tr(2\mathbf{X}\mathbf{A}^\intercal\mathbf{X}^\intercal\mathbf{Q}^\intercal\mathbf{S}^{ij})-tr(2\mathbf{X}\alpha\alpha^\intercal\mathbf{X}^\intercal\mathbf{Q}^\intercal\mathbf{S}^{ij})\nonumber\\+tr
(\beta\Delta \Psi^\intercal\mathbf{S}^{ij})
\end{eqnarray}

By incorporating the gradient of regularization term, i.e., $\Delta\Psi$ into $\eqref{der1H}$ and using Identity number 450 in \cite{matrixcookbook}, we get the following expression:
\begin{eqnarray} \label{derH_2}
\frac{\partial L}{\partial Q_{ij}}=
(2\mathbf{Q}\mathbf{X}\mathbf{A}\mathbf{X}^\intercal)_{ij}-   (2\mathbf{Q}\mathbf{X}\alpha\alpha^\intercal \mathbf{X}^\intercal)_{ij}+\nonumber\\
(\beta2\mathbf{Q} \mathbf{X} \lambda \lambda^\intercal \mathbf{X}^\intercal)_{ij}
\end{eqnarray}.
Now, by taking the second derivative by employing identity 74 from \cite{matrixcookbook}, we obtain the following:
\begin{eqnarray} \label{derH_5}
\frac{\partial}{\partial Q_{kl}}\Big(\frac{\partial L}{\partial Q_{ij}}\Big)=
(2\mathbf{S}^{kl}\mathbf{X}\mathbf{A}\mathbf{X}^\intercal)_{ij}   \nonumber\\
- (2\mathbf{S}^{kl}\mathbf{X}\alpha\alpha^\intercal \mathbf{X}^\intercal)_{ij}
+(\beta2\mathbf{S}^{kl} \mathbf{X} \lambda \lambda^\intercal \mathbf{X}^\intercal)_{ij}.
\end{eqnarray}
We can further express \eqref{derH_5} equivalently as:
\begin{eqnarray} \label{der7Hv2}
\frac{\partial }{\partial Q_{kl}}\Big(\frac{dL}{dQ_{ij}}\Big)=
2tr\big[ \mathbf{X}(\mathbf{A}-\alpha\alpha^\intercal+ \lambda \lambda^\intercal)\mathbf{X}^\intercal(\mathbf{S}^{ij})^\intercal\mathbf{S}^{kl}\big]
\end{eqnarray}
The proof of the equality between \eqref{derH_5} and \eqref{der7Hv2} is provided in the Appendix. 
Utilizing \eqref{der7Hv2}, we can calculate the elements of the Hessian matrix $H_L$. The iterative process of updating the projection matrix, along with the data description, is described in the following sub-section.
\subsection{Subspace optimization and data description}
In order to optimize the subspace along with data description, an iterative two-step process is followed. In the first step, the $\alpha$ values are calculated using SVDD-based optimization; in the second step, the projection matrix $\mathbf{Q}$ is updated using Newton's method-based update. Recently, in \cite{sohrab2023graph}, it is pointed out that the criterion of subspace learning for \ac{OCC} can either be maximized or minimized. Hence, we provide both options for the proposed algorithms to either maximize or minimize the criterion. Algorithm \ref{algo} describes the steps involved in the subspace optimization and data description. 
\begin{algorithm}
\caption{Newton's Method based optimization of subspace support vector data description}\label{algo}
\SetAlgoLined
 \vspace{3mm}
\SetKwInOut{Input}{Input}
\SetKwInOut{Output}{Output}
\Input{$\mathbf{X}$, // Input data\\
$\beta$ // Regularization parameter to control the\\ \quad significance of $\psi$\\
$\eta$, // Learning rate parameter \\
$d$, // Dimensionality of subspace \\
$C$, // Regularization parameter in SVDD\\
min or max // Either minimize or maximize}
\vspace{4mm}
\Output{$\mathbf{Q}$ // Projection matrix \\$R$, // Radius of hypersphere \\$\mbox{\boldmath$\alpha$}$ // Defines the data description }  
 \vspace{4mm}
 // Initialize $\mathbf{Q}$\\
 Random initialization of $\mathbf{Q}$\;
 Orthogonalize $\mathbf{Q}$ using QR decomposition\;
 Row normalize $\mathbf{Q}$ using $l_2$ norm\;
 Initialize $k= 1$\;
 \vspace{3mm}
 \While{$k< k_{max}$ }{
    \vspace{4mm}
    // SVDD in the subspace defined by $\mathbf{Q}$ \\
    Calculate $\mathbf{Y}$ using (\ref{eq:Y_i})\;
    Calculate $\alpha_i,\:i=1,\dots,N$ using (\ref{Lang2})\;
    
    \vspace{4mm}
    // Newton-based update\\
     Calculate $\Delta L$ using (\ref{gradL})\;
     Vectorize $\Delta L$ as $\Delta\mathcal{L}=\text{Vec}(\Delta L)$  using (\ref{vectorizeDeltaL})\;
     Vectorize $\mathbf{Q}$ as $\mathcal{Q}=\text{Vec} (\mathbf{Q})$ using \eqref{vecQ} \\
      Compute the elements of $H_L$ using (\ref{der7Hv2})\;
            \vspace{3mm}   
   \textbf{if} \textit{minimization}\\
 \:\: $\mathcal{Q} \leftarrow \mathcal{Q} - \eta [(H_L)^{-1}\Delta \mathcal{L}]$ \;

  \textbf{elseif} \textit{maximization}\\
 \:\: $\mathcal{Q} \leftarrow \mathcal{Q} + \eta [(H_L)^{-1}\Delta \mathcal{L}]$ \;
    \vspace{3mm}
    // Compute the de-vectorized form of the projection matrix $\mathbf{Q}$ and normalize\\

\For{$i=1:d$}{
\For{$j=1:D$}{
$\mathbf{Q}[i,j]$ = [$\mathcal{Q}$[(i-1)$\times$D + j];\\
}}

    Orthogonalize $\mathbf{Q}$ using QR decomposition\;
    Row normalize $\mathbf{Q}$ using $l_2$ norm\;
   
      $k \leftarrow k+1$ 
   }
   \vspace{4mm}
   
   // SVDD in the optimized subspace\\
   Calculate $\mathbf{Y}$ using (\ref{eq:Y_i})\;
   Calculate $\alpha_i,\:i=1,\dots,N$ using (\ref{Lang2})\;
    Compute the center $\mathbf{a}$ of data description using \eqref{der2}\;
       Identify any support vector $\mathbf{s}$ having $0 < \alpha_s < C$\; 
        \vspace{4mm}
\end{algorithm}

\subsection{nonlinear data description}\label{nonlinear}
For the nonlinear data description, we utilize the nonlinear Projection Trick (NPT) \cite{kwak2013nonlinear}. In the NPT-based approach, we first compute a kernel matrix $\mathbf{K}$ using the radial basis function as
\begin{equation}\label{RBFkernel}
\mathbf{K}_{ij} = \exp  \left( \frac{ -\| \mathbf{x}_{i} - \mathbf{x}_{j}\|_2^2 }{ 2\sigma^2 } \right),
\end{equation} 
where the hyperparameter $\sigma$ scales the distance between the data points $\mathbf{x}_i$ and $\mathbf{x}_j$. The kernel matrix is centered as
\begin{align}\label{centerK}
\mathbf{\Hat{K}} = (\mathbf{I}-  \frac{1}{N}\mathbf{1} \mathbf{1}^\intercal) \mathbf{K} ( \mathbf{I}- \frac{1}{N}\mathbf{1} \mathbf{1}^\intercal),
\end{align}
where $\mathbf{1} \in \mathbb{R}^N$ is a vector with all elements set to one, and the matrix $\mathbf{I}\in\mathbb{R}^{N\times N}$ is an identity matrix. The centered kernel matrix $\mathbf{\Hat{K}}$ is decomposed by using eigendecomposition:
\begin{align}\label{eigen}
\mathbf{\Hat{K}} = \mathbf{U}\mathbf{A}\mathbf{U}^\intercal, 
\end{align}
where $\mathbf{A}$ is a diagonal matrix and contains the non-negative eigenvalues of the matrix $\mathbf{\Hat{K}}$ in its diagonal. The corresponding eigenvectors for the eigenvalues are stored in the columns of matrix $\mathbf{U}$. The data representation for the nonlinear data description is obtained as
\begin{align}\label{nptdata}
\mathbf{\Phi} = (\mathbf{A}^{\frac{1}{2}})^{+} \mathbf{U}^{+} {\mathbf{\Hat{K}} },
\end{align}
where $+$ denotes the pseudo-inverse. After obtaining the data in $\Phi$ space, we apply all the steps used for linear data-description

\subsection{Test phase}\label{SS:Test}
During testing phase, the test instantce $\mathbf{x}_*$ is first mapped to lower $d$-dimensional space as,
\begin{align}\label{test}
\mathbf{y}_* = \mathbf{Q} \mathbf{x}_*.
\end{align}
The decision to classify the test instance as an outlier or target class is taken on the basis of its distance from the center of the data description. The distance is calculated as follows 
\begin{equation}\label{eqtest2}
\|\mathbf{y}_{*} - \mathbf{a}\|_2^2= \mathbf{y}_{*}^\intercal\mathbf{y}_{*} - 2 \sum_{i=1}^{N} \alpha_i \mathbf{y}_{*}^\intercal\mathbf{y}_i + \sum_{i=1}^{N}\sum_{j=1}^{N} \alpha_i \alpha_j \mathbf{y}_i^\intercal\mathbf{y}_j.
\end{equation}
The test instance, $\mathbf{y}_{*}$ in the $d$-dimensional space is classified as positive when $\|\mathbf{y}_{*} - \mathbf{a}\|_2^2 \le R^2$ and as negative, otherwise. During the test phase in the nonlinear case, the kernel vector is computed as
\begin{align}\label{kvector}
\mathbf{k}_{*} = \mathbf{\Phi}^\intercal \phi(\mathbf{x}_{*}).
\end{align}
The kernel vector is then centered as
\begin{align}\label{centerKtest}
\mathbf{{\Hat{k}}}_{*}= (\mathbf{I}- \frac{1}{N}\mathbf{1} \mathbf{1}^\intercal) [  \mathbf{{{k}}}_{*}-\frac{1}{N}\mathbf{K} \mathbf{1}].
\end{align}
The centered kernel vector is then mapped to
\begin{align}\label{npttest}
{\mathbf{\phi}}_{*} = \mathbf{(\Phi}^\intercal)^{+}\mathbf{\Hat{k}}_{*}.
\end{align}
The vector $\mathbf{\phi}_{*}$ is then treated as the test input, and the same steps as those described for the linear test are followed.
\begin{figure*}[ht]
	\centering
	\includegraphics[scale=0.47]{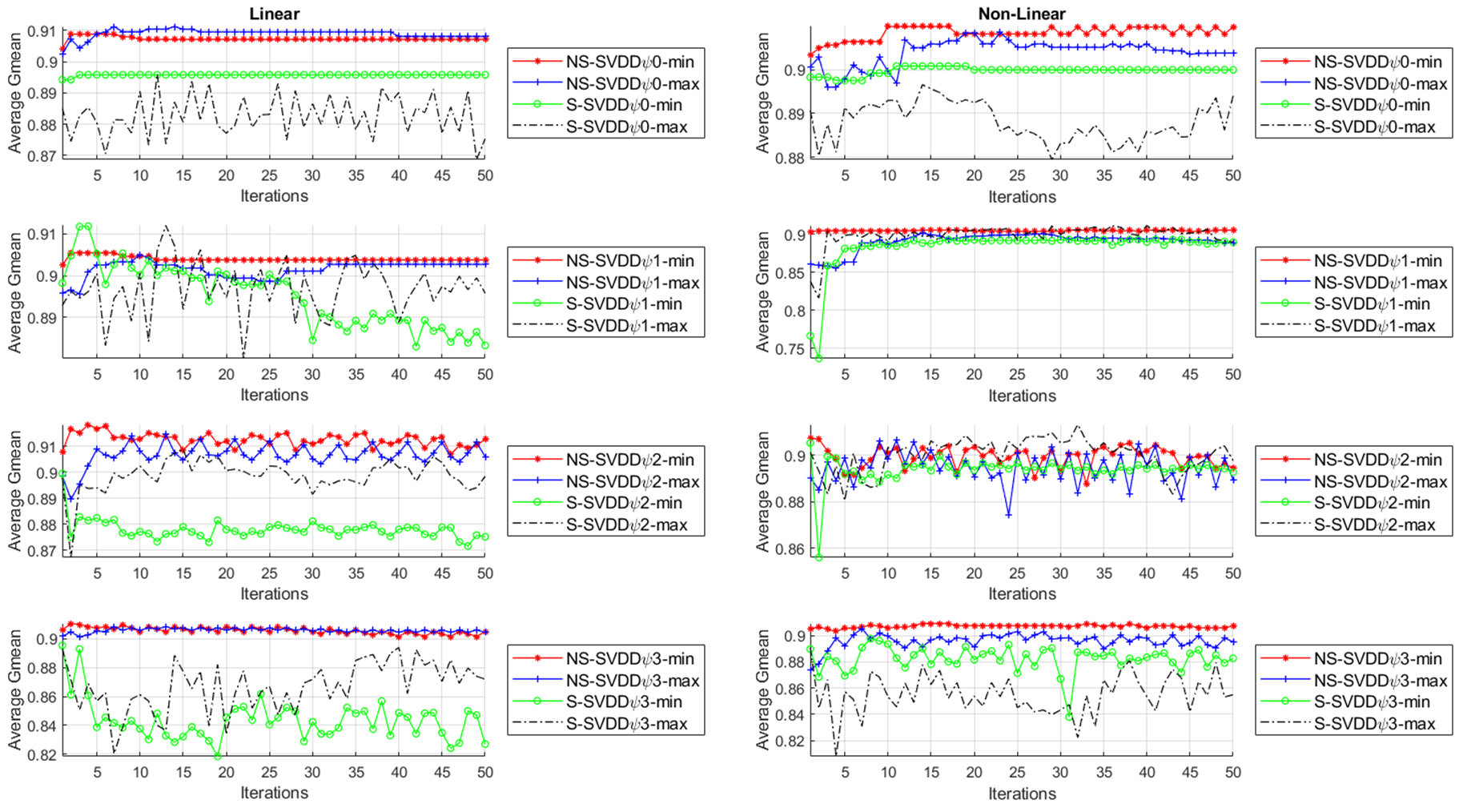}
	\caption{Comparison of different regularization terms for Seeds dataset}
	\label{seeds}
\end{figure*}
\section{Experiments}\label{experiments}
\begin{table*}[ht]
\footnotesize\setlength{\tabcolsep}{8.0pt}
  \centering 
       \caption{\textit{Gmean} results for \textbf{Linear} methods over different datasets}
\begin{tabular}{llllllllllllll}
Dataset                            &                       & \multicolumn{4}{c}{Seeds}                                                                                              & \multicolumn{1}{c}{}  & \multicolumn{3}{c}{Qualitative bankruptcy}                                      & \multicolumn{1}{c}{}  & \multicolumn{3}{c}{Somerville happiness}                                            \\ \cline{1-1} \cline{3-6} \cline{8-10} \cline{12-14} 
\multicolumn{1}{|l|}{Target class} & \multicolumn{1}{l|}{} & \multicolumn{1}{c}{Kama}   & \multicolumn{1}{c}{Rosa}       & \multicolumn{1}{c}{Canadian}  & \multicolumn{1}{c|}{Av.} & \multicolumn{1}{l|}{} & \multicolumn{1}{c}{BK}  & \multicolumn{1}{c}{Non-BK} & \multicolumn{1}{c|}{Av.} & \multicolumn{1}{l|}{} & \multicolumn{1}{c}{Happy} & \multicolumn{1}{c}{Un-happy} & \multicolumn{1}{c|}{Av.} \\ \cline{1-1} \cline{3-6} \cline{8-10} \cline{12-14} 
NS-SVDD$\psi$0-min                      &                       & 0.83                       & 0.91                           & 0.88                          & 0.88                     &                       & 0.91                    & 0.05                       & 0.48                     &                       & 0.40                      & 0.47                         & 0.44                     \\
NS-SVDD$\psi$1-min                      &                       & 0.85                       & 0.90                           & 0.93                          & 0.90                     &                       & 0.90                    & 0.03                       & 0.47                     &                       & 0.45                      & 0.46                         & 0.45                     \\
NS-SVDD$\psi$2-min                      &                       & 0.83                       & 0.93                           & 0.93                          & 0.90                     &                       & 0.85                    & 0.09                       & 0.47                     &                       & 0.47                      & 0.35                         & 0.41                     \\
NS-SVDD$\psi$3-min                      &                       & 0.82                       & 0.91                           & 0.88                          & 0.87                     &                       & 0.90                    & 0.21                       & 0.55                     &                       & 0.46                      & 0.41                         & 0.43                     \\
NS-SVDD$\psi$0-max                      &                       & \textbf{0.85}              & 0.90                           & 0.94                          & 0.90                     &                       & 0.87                    & 0.00                       & 0.44                     &                       & 0.42                      & 0.48                         & 0.45                     \\
NS-SVDD$\psi$1-max                      &                       & 0.82                       & 0.91                           & 0.93                          & 0.88                     &                       & 0.90                    & 0.00                       & 0.45                     &                       & 0.38                      & 0.46                         & 0.42                     \\
NS-SVDD$\psi$2-max                      &                       & \textbf{0.85}              & 0.89                           & 0.89                          & 0.88                     &                       & 0.91                    & 0.16                       & 0.54                     &                       & 0.48                      & 0.41                         & 0.44                     \\
NS-SVDD$\psi$3-max                      &                       & 0.81                       & 0.90                           & 0.77                          & 0.83                     &                       & 0.87                    & 0.19                       & 0.53                     &                       & 0.39                      & 0.43                         & 0.41                     \\
S-SVDD$\psi$0-min                      &                       & 0.85                       & 0.92                           & \textbf{0.95}                 & \textbf{0.91}            &                       & 0.91                    & 0.21                       & 0.56                     &                       & 0.47                      & 0.41                         & 0.44                     \\
S-SVDD$\psi$1-min                      &                       & 0.84                       & 0.93                           & 0.94                          & 0.90                     &                       & 0.82                    & 0.24                       & 0.53                     &                       & \textbf{0.51}             & 0.41                         & 0.46                     \\
S-SVDD$\psi$2-min                      &                       & 0.84                       & 0.94                           & 0.94                          & \textbf{0.91}            &                       & 0.93                    & 0.16                       & 0.55                     &                       & 0.49                      & 0.43                         & 0.46                     \\
S-SVDD$\psi$3-min                      &                       & \textbf{0.85}              & 0.91                           & 0.95                          & 0.90                     &                       & 0.92                    & 0.07                       & 0.49                     &                       & 0.51                      & 0.44                         & 0.48                     \\
S-SVDD$\psi$0-max                      &                       & \textbf{0.85}              & 0.93                           & 0.94                          & \textbf{0.91}            &                       & 0.86                    & 0.00                       & 0.43                     &                       & 0.40                      & 0.44                         & 0.42                     \\
S-SVDD$\psi$1-max                      &                       & \textbf{0.85}              & 0.92                           & 0.94                          & 0.90                     &                       & 0.86                    & 0.00                       & 0.43                     &                       & 0.42                      & 0.45                         & 0.44                     \\
S-SVDD$\psi$2-max                      &                       & \textbf{0.85}              & \textbf{0.95}                  & 0.94                          & \textbf{0.91}            &                       & 0.88                    & 0.03                       & 0.45                     &                       & 0.37                      & 0.43                         & 0.40                     \\
S-SVDD$\psi$3-max                      &                       & 0.84                       & 0.93                           & 0.94                          & 0.90                     &                       & 0.88                    & 0.00                       & 0.44                     &                       & 0.41                      & 0.45                         & 0.43                     \\
E-SVDD                              &                       & 0.79                       & 0.87                           & 0.87                          & 0.84                     &                       & \textbf{0.96}           & 0.19                       & \textbf{0.58}            &                       & 0.42                      & 0.41                         & 0.41                     \\
SVDD                               &                       & \textbf{0.85}              & 0.92                           & 0.94                          & 0.90                     &                       & 0.94                    & 0.00                       & 0.47                     &                       & 0.41                      & 0.36                         & 0.39                     \\
OC-SVM                              &                       & 0.48                       & 0.69                           & 0.45                          & 0.54                     &                       & 0.37                    & \textbf{0.41}              & 0.39                     &                       & 0.45                      & \textbf{0.53}                & \textbf{0.49}            \\
                                   &                       &                            &                                &                               &                          &                       &                         &                            &                          &                       &                           &                              &                          \\
Dataset                            &                       & \multicolumn{4}{c}{Iris}                                                                                               & \multicolumn{1}{c}{}  & \multicolumn{3}{c}{Ionosphere}                                                  & \multicolumn{1}{c}{}  & \multicolumn{3}{c}{Sonar}                                                           \\ \cline{1-1} \cline{3-6} \cline{8-10} \cline{12-14} 
\multicolumn{1}{|l|}{Target class} & \multicolumn{1}{l|}{} & \multicolumn{1}{c}{Setosa} & \multicolumn{1}{c}{Versicolor} & \multicolumn{1}{c}{Virginica} & \multicolumn{1}{c|}{Av.} & \multicolumn{1}{l|}{} & \multicolumn{1}{c}{Bad} & \multicolumn{1}{c}{Good}   & \multicolumn{1}{c|}{Av.} & \multicolumn{1}{l|}{} & \multicolumn{1}{c}{Rock}  & \multicolumn{1}{c}{Mines}    & \multicolumn{1}{c|}{Av.} \\ \cline{1-1} \cline{3-6} \cline{8-10} \cline{12-14} 
NS-SVDD$\psi$0-min                      &                       & 0.93                       & 0.91                           & 0.87                          & 0.90                     &                       & 0.37                    & 0.91                       & 0.64                     &                       & 0.54                      & 0.61                         & \textbf{0.58}            \\
NS-SVDD$\psi$1-min                      &                       & 0.94                       & 0.90                           & 0.82                          & 0.89                     &                       & 0.46                    & \textbf{0.92}              & \textbf{0.69}            &                       & 0.55                      & 0.61                         & \textbf{0.58}            \\
NS-SVDD$\psi$2-min                      &                       & 0.94                       & 0.92                           & 0.87                          & 0.91                     &                       & 0.39                    & 0.91                       & 0.65                     &                       & 0.51                      & \textbf{0.64}                & \textbf{0.58}            \\
NS-SVDD$\psi$3-min                      &                       & 0.94                       & 0.88                           & 0.87                          & 0.89                     &                       & 0.27                    & \textbf{0.92}              & 0.60                     &                       & 0.51                      & 0.58                         & 0.55                     \\
NS-SVDD$\psi$0-max                      &                       & 0.94                       & 0.90                           & 0.87                          & 0.90                     &                       & 0.05                    & 0.79                       & 0.42                     &                       & 0.49                      & 0.40                         & 0.45                     \\
NS-SVDD$\psi$1-max                      &                       & 0.96                       & 0.89                           & 0.89                          & 0.91                     &                       & 0.25                    & 0.91                       & 0.58                     &                       & \textbf{0.55}             & 0.44                         & 0.50                     \\
NS-SVDD$\psi$2-max                      &                       & 0.96                       & \textbf{0.94}                  & 0.86                          & 0.92                     &                       & 0.13                    & 0.82                       & 0.47                     &                       & 0.38                      & 0.45                         & 0.41                     \\
NS-SVDD$\psi$3-max                      &                       & 0.90                       & 0.91                           & 0.89                          & 0.90                     &                       & 0.18                    & 0.87                       & 0.52                     &                       & 0.40                      & 0.41                         & 0.40                     \\
S-SVDD$\psi$0-min                      &                       & 0.94                       & 0.92                           & \textbf{0.91}                 & \textbf{0.92}            &                       & 0.16                    & 0.79                       & 0.48                     &                       & 0.50                      & 0.52                         & 0.51                     \\
S-SVDD$\psi$1-min                      &                       & 0.94                       & 0.92                           & 0.88                          & 0.91                     &                       & 0.16                    & 0.81                       & 0.49                     &                       & 0.48                      & 0.54                         & 0.51                     \\
S-SVDD$\psi$2-min                      &                       & \textbf{0.97}              & 0.91                           & 0.88                          & 0.92                     &                       & 0.19                    & 0.81                       & 0.50                     &                       & 0.52                      & 0.55                         & 0.53                     \\
S-SVDD$\psi$3-min                      &                       & 0.95                       & 0.93                           & 0.89                          & 0.92                     &                       & 0.19                    & 0.79                       & 0.49                     &                       & 0.50                      & 0.55                         & 0.52                     \\
S-SVDD$\psi$0-max                      &                       & 0.94                       & 0.91                           & 0.89                          & 0.91                     &                       & 0.12                    & 0.79                       & 0.46                     &                       & 0.49                      & 0.41                         & 0.45                     \\
S-SVDD$\psi$1-max                      &                       & 0.94                       & 0.91                           & 0.87                          & 0.91                     &                       & 0.12                    & 0.81                       & 0.46                     &                       & 0.48                      & 0.54                         & 0.51                     \\
S-SVDD$\psi$2-max                      &                       & 0.92                       & 0.84                           & 0.89                          & 0.89                     &                       & 0.15                    & 0.79                       & 0.47                     &                       & 0.50                      & 0.41                         & 0.45                     \\
S-SVDD$\psi$3-max                      &                       & 0.93                       & 0.90                           & 0.89                          & 0.91                     &                       & 0.12                    & 0.79                       & 0.46                     &                       & 0.49                      & 0.41                         & 0.45                     \\
E-SVDD                              &                       & 0.89                       & 0.85                           & 0.86                          & 0.87                     &                       & 0.33                    & 0.88                       & 0.61                     &                       & 0.00                      & 0.03                         & 0.02                     \\
SVDD                               &                       & 0.92                       & 0.90                           & 0.89                          & 0.91                     &                       & 0.02                    & 0.86                       & 0.44                     &                       & 0.52                      & 0.56                         & 0.54                     \\
OC-SVM                              &                       & 0.58                       & 0.50                           & 0.46                          & 0.51                     &                       & \textbf{0.49}           & 0.51                       & 0.50                     &                       & 0.48                      & 0.45                         & 0.46                    
\end{tabular}
\label{linearresults}  
\end{table*}
\begin{table*}[ht]
\footnotesize\setlength{\tabcolsep}{8.0pt}
  \centering 
       \caption{\textit{Gmean} results for \textbf{nonlinear} methods over different datasets}
\begin{tabular}{llcccclccclccc}
Dataset                            &                       & \multicolumn{4}{c}{Seeds}                                                                     & \multicolumn{1}{c}{}  & \multicolumn{3}{c}{Qualitative bankruptcy}                             & \multicolumn{1}{c}{}  & \multicolumn{3}{c}{Somerville happiness}                               \\ \cline{1-1} \cline{3-6} \cline{8-10} \cline{12-14} 
\multicolumn{1}{|l|}{Target class} & \multicolumn{1}{l|}{} & Kama                 & Rosa                 & Canadian             & \multicolumn{1}{c|}{Av.} & \multicolumn{1}{l|}{} & BK                   & Non-BK               & \multicolumn{1}{c|}{Av.} & \multicolumn{1}{l|}{} & Happy                & Un-happy             & \multicolumn{1}{c|}{Av.} \\ \cline{1-1} \cline{3-6} \cline{8-10} \cline{12-14} 
NS-SVDD$\psi$0-min                      &                       & 0.84                 & 0.91                 & 0.94                 & 0.89                     &                       & 0.87                 & 0.49                 & 0.68                     &                       & 0.51                 & 0.42                 & 0.47                     \\
NS-SVDD$\psi$1-min                      &                       & 0.81                 & 0.92                 & 0.94                 & 0.89                     &                       & 0.93                 & \textbf{0.62}        & \textbf{0.77}            &                       & 0.51                 & 0.50                 & 0.50                     \\
NS-SVDD$\psi$2-min                      &                       & 0.83                 & \textbf{0.95}        & 0.91                 & 0.90                     &                       & 0.93                 & 0.58                 & 0.76                     &                       & 0.49                 & 0.45                 & 0.47                     \\
NS-SVDD$\psi$3-min                      &                       & 0.82                 & 0.89                 & 0.85                 & 0.85                     &                       & 0.91                 & 0.46                 & 0.69                     &                       & \textbf{0.57}        & 0.42                 & 0.50                     \\
NS-SVDD$\psi$0-max                      &                       & \textbf{0.85}        & 0.92                 & 0.94                 & 0.90                     &                       & 0.94                 & 0.50                 & 0.72                     &                       & 0.43                 & 0.46                 & 0.44                     \\
NS-SVDD$\psi$1-max                      &                       & 0.83                 & 0.91                 & 0.93                 & 0.89                     &                       & 0.94                 & 0.49                 & 0.71                     &                       & 0.55                 & 0.46                 & \textbf{0.51}            \\
NS-SVDD$\psi$2-max                      &                       & 0.83                 & 0.91                 & 0.94                 & 0.89                     &                       & 0.94                 & 0.49                 & 0.71                     &                       & 0.42                 & 0.31                 & 0.37                     \\
NS-SVDD$\psi$3-max                      &                       & 0.81                 & 0.92                 & 0.92                 & 0.88                     &                       & 0.93                 & 0.48                 & 0.70                     &                       & 0.47                 & 0.46                 & 0.47                     \\
S-SVDD$\psi$0-min                      &                       & 0.83                 & 0.94                 & 0.94                 & 0.90                     &                       & 0.91                 & 0.47                 & 0.69                     &                       & 0.48                 & 0.37                 & 0.43                     \\
S-SVDD$\psi$1-min                      &                       & \textbf{0.85}        & 0.90                 & 0.91                 & 0.89                     &                       & 0.92                 & 0.53                 & 0.73                     &                       & 0.34                 & \textbf{0.52}        & 0.43                     \\
S-SVDD$\psi$2-min                      &                       & 0.82                 & 0.91                 & 0.94                 & 0.89                     &                       & 0.93                 & 0.46                 & 0.69                     &                       & 0.47                 & 0.36                 & 0.42                     \\
S-SVDD$\psi$3-min                      &                       & 0.82                 & 0.93                 & 0.94                 & 0.90                     &                       & 0.93                 & 0.43                 & 0.68                     &                       & 0.51                 & 0.39                 & 0.45                     \\
S-SVDD$\psi$0-max                      &                       & \textbf{0.85}        & 0.94                 & 0.94                 & \textbf{0.91}            &                       & 0.94                 & 0.51                 & 0.72                     &                       & 0.44                 & 0.47                 & 0.46                     \\
S-SVDD$\psi$1-max                      &                       & \textbf{0.85}        & 0.93                 & 0.94                 & \textbf{0.91}            &                       & 0.94                 & 0.51                 & 0.72                     &                       & 0.43                 & 0.48                 & 0.46                     \\
S-SVDD$\psi$2-max                      &                       & 0.84                 & 0.91                 & 0.94                 & 0.89                     &                       & 0.90                 & 0.39                 & 0.65                     &                       & 0.40                 & 0.39                 & 0.40                     \\
S-SVDD$\psi$3-max                      &                       & 0.84                 & 0.94                 & 0.94                 & \textbf{0.91}            &                       & 0.93                 & 0.51                 & 0.72                     &                       & 0.44                 & 0.46                 & 0.45                     \\
E-SVDD                              &                       & 0.81                 & 0.88                 & 0.87                 & 0.85                     &                       & 0.00                 & 0.00                 & 0.00                     &                       & 0.00                 & 0.31                 & 0.16                     \\
SVDD                               &                       & \textbf{0.85}        & 0.91                 & \textbf{0.95}        & 0.90                     &                       & 0.33                 & 0.28                 & 0.31                     &                       & 0.40                 & 0.32                 & 0.36                     \\
OC-SVM                              &                       & 0.47                 & 0.60                 & 0.45                 & 0.51                     &                       & 0.36                 & 0.58                 & 0.47                     &                       & 0.47                 & 0.49                 & 0.48                     \\
GE-SVDD                        &                       & 0.85                 & 0.93                 & 0.93                 & 0.90                     &                       & 0.94                 & 0.28                 & 0.61                     &                       & 0.50                 & 0.48                 & 0.49                     \\
GE-SVM                         &                       & \textbf{0.85}        & 0.90                 & 0.93                 & 0.89                     &                       & \textbf{0.95}        & 0.26                 & 0.60                     &                       & 0.52                 & 0.48                 & 0.50                     \\
                                   &                       & \multicolumn{1}{l}{} & \multicolumn{1}{l}{} & \multicolumn{1}{l}{} & \multicolumn{1}{l}{}     &                       & \multicolumn{1}{l}{} & \multicolumn{1}{l}{} & \multicolumn{1}{l}{}     &                       & \multicolumn{1}{l}{} & \multicolumn{1}{l}{} & \multicolumn{1}{l}{}     \\
Dataset                            &                       & \multicolumn{4}{c}{Iris}                                                                      & \multicolumn{1}{c}{}  & \multicolumn{3}{c}{Ionosphere}                                         & \multicolumn{1}{c}{}  & \multicolumn{3}{c}{Sonar}                                              \\ \cline{1-1} \cline{3-6} \cline{8-10} \cline{12-14} 
\multicolumn{1}{|l|}{Target class} & \multicolumn{1}{l|}{} & Setosa               & Versicolor           & Virginica            & \multicolumn{1}{c|}{Av.} & \multicolumn{1}{l|}{} & Bad                  & Good                 & \multicolumn{1}{c|}{Av.} & \multicolumn{1}{l|}{} & Rock                 & Mines                & \multicolumn{1}{c|}{Av.} \\ \cline{1-1} \cline{3-6} \cline{8-10} \cline{12-14} 
NS-SVDD$\psi$0-min                      &                       & 0.93                 & 0.82                 & 0.90                 & 0.88                     &                       & 0.50                 & 0.90                 & 0.70                     &                       & \textbf{0.59}        & \textbf{0.64}        & \textbf{0.62}            \\
NS-SVDD$\psi$1-min                      &                       & 0.93                 & \textbf{0.93}        & 0.85                 & 0.90                     &                       & 0.63                 & 0.90                 & 0.77                     &                       & 0.57                 & 0.56                 & 0.56                     \\
NS-SVDD$\psi$2-min                      &                       & 0.94                 & 0.92                 & 0.84                 & 0.90                     &                       & 0.68                 & 0.90                 & \textbf{0.79}            &                       & 0.50                 & 0.58                 & 0.54                     \\
NS-SVDD$\psi$3-min                      &                       & 0.89                 & 0.83                 & \textbf{0.93}        & 0.88                     &                       & 0.59                 & 0.89                 & 0.74                     &                       & 0.56                 & 0.60                 & 0.58                     \\
NS-SVDD$\psi$0-max                      &                       & 0.94                 & 0.91                 & 0.88                 & 0.91                     &                       & 0.32                 & 0.78                 & 0.55                     &                       & 0.49                 & 0.47                 & 0.48                     \\
NS-SVDD$\psi$1-max                      &                       & 0.96                 & 0.90                 & 0.81                 & 0.89                     &                       & 0.67                 & 0.91                 & 0.79                     &                       & 0.53                 & 0.59                 & 0.56                     \\
NS-SVDD$\psi$2-max                      &                       & 0.88                 & 0.91                 & 0.91                 & 0.90                     &                       & \textbf{0.70}        & 0.81                 & 0.76                     &                       & 0.45                 & 0.48                 & 0.46                     \\
NS-SVDD$\psi$3-max                      &                       & 0.94                 & 0.88                 & 0.58                 & 0.80                     &                       & 0.57                 & 0.80                 & 0.68                     &                       & 0.49                 & 0.45                 & 0.47                     \\
S-SVDD$\psi$0-min                      &                       & 0.94                 & 0.90                 & 0.90                 & 0.91                     &                       & 0.51                 & 0.90                 & 0.70                     &                       & 0.53                 & 0.51                 & 0.52                     \\
S-SVDD$\psi$1-min                      &                       & 0.96                 & 0.88                 & 0.90                 & 0.91                     &                       & 0.46                 & \textbf{0.92}        & 0.69                     &                       & 0.53                 & 0.54                 & 0.54                     \\
S-SVDD$\psi$2-min                      &                       & 0.85                 & 0.91                 & 0.91                 & 0.89                     &                       & 0.64                 & 0.90                 & 0.77                     &                       & 0.56                 & 0.56                 & 0.56                     \\
S-SVDD$\psi$3-min                      &                       & 0.92                 & 0.91                 & 0.90                 & 0.91                     &                       & 0.51                 & 0.91                 & 0.71                     &                       & 0.58                 & 0.48                 & 0.53                     \\
S-SVDD$\psi$0-max                      &                       & \textbf{0.96}        & 0.91                 & 0.89                 & \textbf{0.92}            &                       & 0.39                 & 0.80                 & 0.60                     &                       & 0.48                 & 0.53                 & 0.50                     \\
S-SVDD$\psi$1-max                      &                       & \textbf{0.96}        & 0.91                 & 0.87                 & 0.91                     &                       & 0.41                 & 0.80                 & 0.60                     &                       & 0.49                 & 0.58                 & 0.54                     \\
S-SVDD$\psi$2-max                      &                       & 0.94                 & 0.92                 & 0.88                 & 0.91                     &                       & 0.42                 & 0.82                 & 0.62                     &                       & 0.47                 & 0.48                 & 0.48                     \\
S-SVDD$\psi$3-max                      &                       & 0.96                 & 0.89                 & 0.90                 & 0.92                     &                       & 0.41                 & 0.82                 & 0.61                     &                       & 0.48                 & 0.53                 & 0.50                     \\
E-SVDD                              &                       & 0.68                 & 0.84                 & 0.83                 & 0.78                     &                       & 0.37                 & 0.88                 & 0.63                     &                       & 0.55                 & 0.52                 & 0.53                     \\
SVDD                               &                       & 0.92                 & 0.92                 & 0.88                 & 0.90                     &                       & 0.21                 & 0.85                 & 0.53                     &                       & 0.53                 & 0.59                 & 0.56                     \\
OC-SVM                              &                       & 0.56                 & 0.26                 & 0.55                 & 0.46                     &                       & 0.52                 & 0.47                 & 0.49                     &                       & 0.47                 & 0.55                 & 0.51                     \\
GE-SVDD                        &                       & 0.83                 & 0.92                 & 0.89                 & 0.88                     &                       & 0.38                 & 0.88                 & 0.63                     &                       & 0.55                 & 0.60                 & 0.57                     \\
GE-SVM                         &                       & 0.90                 & 0.90                 & 0.90                 & 0.90                     &                       & 0.38                 & 0.91                 & 0.64                     &                       & 0.52                 & 0.61                 & 0.57                    
\end{tabular}
\label{nonlinearresults}  
\end{table*}

\subsection{Datasets, evaluation criteria and experimental setup}
For both linear and nonlinear data descriptions, we evaluated the algorithms over six different datasets downloaded from the UCI machine learning repository \cite{Dua:2019}. The datasets used in the experiments are Seeds, Qualitative bankruptcy, Somerville happiness, Iris, Ionosphere, and Sonar. The Seeds dataset consists of three classes (Kama, Rosa, Canadian), with a total of 210 samples and 70 target samples per class and a dimensionality of 7. The Qualitative bankruptcy dataset has two classes, bankruptcy (BK) and non-bankruptcy (Non-BK), with 250 total samples. The BK class has 107 target samples, the Non-BK class has 143 target samples, and the dimensionality is 6. The Somerville happiness dataset has two classes (Happy, and Un-happy) with 143 total samples. The happy class has 77 target samples, the Un-happy class has 66 target samples, and the dimensionality is also 6. The Iris dataset has three classes (Setosa, Versicolor, Virginica) with 150 samples in total and 50 target samples per class. The dimensionality for this dataset is 4. The Ionosphere dataset consists of two classes (Bad, Good) with 351 total samples. The bad class has 126 target samples, the good class has 225 target samples, and the dimensionality is 34. Finally, the Sonar dataset contains two classes (Rock, Mines) with 208 total samples. The rock class has 97 target samples, the mines class has 111 target samples, and the dimensionality is 60. We designated a single class at a time as the target class and the rest as outliers.

For each dataset, we randomly divided the data into 70\% for training and the remaining 30\% for testing, with the proportions of classes similar to the complete dataset. We repeated each experiment five times, using different train/test splits. We report the average test performance over the five splits. During training, we utilized 5-fold cross-validation for selecting the optimal hyperparameters based on the best Geometric mean (\textit{Gmean}) score. \textit{Gmean} is defined as
\begin{equation}\label{gmean}
 Gmean=\sqrt {tpr \times tnr},
\end{equation}
where $tpr$ is the true positive rate, and $tnr$ is the true negative rate. We chose the hyperparameters from the following range of values:
\begin{itemize}
  \item$\beta\in\{10^{-2},10^{-1},10^{0},10^{1},10^{2}\},$
  \item$C\in\{0.01,0.05,0.1,0.2,0.3,0.4,0.5\}$,
  \item$\sigma\in\{10^{-1},10^{0},10^{1},10^{2},10^{3}\}$,
  \item$d\in\{1,2,3,4,5,10,20\}$,
  \item$\eta\in\{10^{-5},10^{-4},10^{-3},10^{-2},10^{-1}\}$.
\end{itemize}

\subsection{Experimental results and discussion}
In our evaluation, we compared the proposed NS-SVDD method with its counterpart, S-SVDD, as well as several other \ac{OCC} methods. These methods include SVDD, Ellipsoidal-SVDD (E-SVDD), OC-SVM, Graph-Embedded SVM (GE-SVM), and Graph-Embedded SVDD (GE-SVDD). The performance of the proposed NS-SVDD method, along with other competing methods, is presented in Tables \ref{linearresults} and \ref{nonlinearresults} for linear and nonlinear data description, respectively. The average performance across each dataset is reported in the "average" (Av.) column. Additionally, for NS-SVDD and S-SVDD, it is possible to optimize the criterion either by minimizing (-min) or maximizing (-max) it. Hence, we conducted separate experiments to maximize and minimize the overall criterion for different variants and compare the results. 

The maximization strategy demonstrated that the proposed Newton-based solution outperformed the gradient-based solution in the majority of cases, both for linear and nonlinear data description. Regarding the minimization strategy, the gradient-based solution performed better in the linear case, while the Newton-based method showed superior performance in the nonlinear case.

Overall, the Newton-based method in the minimization strategy exhibited better results compared to the Newton-based method in the maximization strategy in most cases, irrespective of linear or nonlinear data description. Similarly, in the gradient-based approach, the minimization strategy yielded better outcomes in the linear case and performed equally well as the maximization strategy in the nonlinear case.

We evaluate and present the performance of the proposed NS-SVDD and the gradient-based S-SVDD methods on the test set at each training iteration for both linear and nonlinear scenarios. The average \textit{Gmean} value is computed for each iteration across the five test splits for the Seeds dataset, which can be seen in Figure \ref{seeds}. We report the performance of these methods under various settings of the regularization term $\psi$. To compare their performance, we utilize the \textit{Gmean} as an evaluating metric. Moreover, we report the results for both maximizing and minimizing the criterion to provide a comprehensive analysis. Similar figures can be generated for all other datasets as well. We notice that the proposed Newton-based NS-SVDD method demonstrated more stability and faster convergence to its optimal performance compared to the gradient-based S-SVDD method, as evidenced by the plotted results across different regularization terms. Considering the regularization strategies ($\psi$0-$\psi3$), the regularization strategy $\psi$2 was found to be effective in the linear case, while the regularization strategy $\psi$1 showed promising results in the nonlinear case.

\section{CONCLUSION}\label{Conclusions}

In this paper, we propose a novel NS-SVDD which leverages Newton’s method to
enhance data mapping and data description for \ac{OCC}. We defined both linear and nonlinear versions for the proposed Newton-based optimization of subspace learning for \ac{OCC}. We experimented with both minimization and maximization strategies and evaluated the algorithms with different regularization terms. Our findings indicate that the Newton-based method outperformed the gradient-based method in most cases for both linear and nonlinear data description in the maximization strategy. However, in the minimization strategy, the gradient-based method performed better for linear cases, while the Newton-based method performed better for nonlinear cases. Overall, the Newton-based method with the minimization strategy performed better than the maximization strategy for both linear and nonlinear data descriptions. In the future, we will investigate the usage of Newton's method in the graph-embedded based subspace learning methods for one-class classification and extend it to multi-modal \ac{OCC} \cite{sohrab2021multimodal} as well.

\section*{Acknowledgement}
This work was supported by the NSF-Business Finland project AMALIA. Foundation for Economic
Education (Grant number: 220363) funded the work of Fahad Sohrab at Haltian.

\section*{APPENDIX}
Proof that \eqref{derH_5} can be written in the form of \eqref{der7Hv2}. We start by taking the derivate of \eqref{der1H}. 
\begin{align} \label{der2H}
\frac{\partial}{\partial Q_{kl}}\Big(\frac{\partial L}{\partial Q_{ij}}\Big) & =
\frac{\partial}{dQ_{kl}}tr(2\mathbf{X}\mathbf{A}^\intercal\mathbf{X}^\intercal\mathbf{Q}^\intercal\mathbf{S}^{ij})\nonumber \nonumber\\ 
& -\frac{\partial}{dQ_{kl}}tr(2\mathbf{X}\alpha\alpha^\intercal\mathbf{X}^\intercal\mathbf{Q}^\intercal\mathbf{S}^{ij})\nonumber\\ 
& +\frac{\partial}{dQ_{kl}}tr
(2\beta\mathbf{X} \lambda \lambda^\intercal \mathbf{X}^\intercal  \mathbf{Q}^\intercal\mathbf{S}^{ij})\nonumber \\
&=tr\Big[ \big[\frac{\partial }{\partial \mathbf{Q}}tr(2\mathbf{X}\mathbf{A}^\intercal\mathbf{X}^\intercal\mathbf{Q}^\intercal\mathbf{S}^{ij})\big]^\intercal  \frac{\partial\mathbf{Q}}{dQ_{kl}} \Big]
\nonumber\\
&-tr\Big[\big[\frac{\partial}{\partial\mathbf{Q}}tr(2\mathbf{X}\alpha\alpha^\intercal\mathbf{X}^\intercal\mathbf{Q}^\intercal\mathbf{S}^{ij})\big]^\intercal\frac{\partial\mathbf{Q}}{\partial Q_{kl}}\Big]\nonumber\\
&+tr\Big[\big[\frac{\partial}{\partial\mathbf{Q}}tr
(2\beta\mathbf{X} \lambda \lambda^\intercal \mathbf{X}^\intercal  \mathbf{Q}^\intercal\mathbf{S}^{ij})\big]^\intercal\frac{\partial\mathbf{Q}}{dQ_{kl}}\Big] \nonumber\\
&=
tr\Big[ \big[2\mathbf{S}^{ij}\mathbf{X}\mathbf{A}^\intercal\mathbf{X}^\intercal\big]^\intercal  \mathbf{S}^{kl} \Big]
\nonumber\\ &-
tr\Big[\big[2\mathbf{S}^{ij}\mathbf{X}\alpha\alpha^\intercal\mathbf{X}^\intercal\big]^\intercal\mathbf{S}^{kl}\Big] \nonumber\\
&+tr\Big[\big[2\beta\mathbf{S}^{ij}  \mathbf{X} \lambda \lambda^\intercal \mathbf{X}^\intercal \big]^\intercal\mathbf{S}^{kl}\Big] \nonumber \\ 
&=
2tr\Big[ \mathbf{X}\mathbf{A}\mathbf{X}^\intercal(\mathbf{S}^{ij})^\intercal \mathbf{S}^{kl} \Big]
\nonumber\\ 
&- 2tr \Big[\mathbf{X}\alpha\alpha^\intercal\mathbf{X}^\intercal(\mathbf{S}^{ij})^\intercal\mathbf{S}^{kl}\Big] \nonumber\\
& +2\beta tr\Big[  \mathbf{X} \lambda \lambda^\intercal \mathbf{X}^\intercal (\mathbf{S}^{ij})^\intercal\mathbf{S}^{kl}\Big] \nonumber\\
& =
2tr\Big[ \mathbf{X}\mathbf{A}\mathbf{X}^\intercal(\mathbf{S}^{ij})^\intercal \mathbf{S}^{kl} 
\nonumber\\
& -
\mathbf{X}\alpha\alpha^\intercal\mathbf{X}^\intercal(\mathbf{S}^{ij})^\intercal\mathbf{S}^{kl}
\nonumber\\
&  +\beta  \mathbf{X} \lambda \lambda^\intercal \mathbf{X}^\intercal (\mathbf{S}^{ij})^\intercal\mathbf{S}^{kl}\Big] \nonumber\\
& 
=
2tr\big[ \mathbf{X}(\mathbf{A}-\alpha\alpha^\intercal+ \lambda \lambda^\intercal)\mathbf{X}^\intercal(\mathbf{S}^{ij})^\intercal\mathbf{S}^{kl}\big] \nonumber
\end{align}

\end{document}